\documentclass[twocolumn, switch]{article} 

\usepackage{preprint}

\usepackage[utf8]{inputenc}	
\usepackage[T1]{fontenc}	
\usepackage{xcolor}		
\usepackage[colorlinks = false,
            linkcolor = purple,
            urlcolor  = blue,
            citecolor = cyan,
            anchorcolor = black]{hyperref}	

\usepackage{amsmath,amssymb,amsfonts}
\usepackage{algorithmic}
\usepackage{graphicx}
\usepackage{textcomp}
\usepackage{color}
\usepackage{xcolor}
\usepackage{amsmath}
\usepackage{comment}
\usepackage{pifont}
\usepackage{bm,upgreek}
\usepackage{mathrsfs}
\usepackage{url}
\usepackage{colortbl}
\usepackage{svg}
\usepackage{color}
\usepackage{booktabs}
\usepackage{graphicx}
\usepackage{import}
\usepackage{wrapfig}
\usepackage{diagbox}
\usepackage{subcaption}
\usepackage{multicol}
\usepackage{tablefootnote}
\usepackage{multirow}
\usepackage{wrapfig}

\usepackage[linesnumbered, ruled, vlined]{algorithm2e}

\newcommand{\hati}{\hat{\textbf{ı}}}
\newcommand{\hatj}{\hat{\mathbf{j}}}
\newcommand{\hatk}{\hat{\mathbf{k}}}
\newcommand{\uvec}[1]{\hat{\mathbf{#1}}}
\newcommand{\pv}[3]{{}^{#1}\mathbf{#2}_{#3}}
\newcommand{\uv}[3]{{}^{#1}\uvec{#2}_{#3}}
\newcommand{\ui}[2]{{}^{#1}\hati_{#2}}
\newcommand{\uj}[2]{{}^{#1}\hatj_{#2}}
\newcommand{\uk}[2]{{}^{#1}\hatk_{#2}}
\newcommand{\rotmat}[2]{{}^{#1}_{#2}\mathbf{R}}

\newcommand{\wcolzer}{0.1cm}
\newcommand{\wcolone}{0.7cm}
\newcommand{\wcoltwo}{0.7cm}
\newcommand{\wcolthr}{0.8cm}
\newcommand{\wcolfou}{0.8cm}
\newcommand{\wcolfiv}{0.5cm}
\newcommand{\wcolsix}{0.7cm}
\newcommand{\wcolsev}{0.8cm}

\usepackage{tikz}    

\title{GeoFIK: A Fast and Reliable Geometric Solver for the IK of the Franka Arm Based on Screw Theory Enabling Multiple Redundancy Parameters}

\usepackage{authblk}

\author[1\thanks{\tt{custodio825@gmail.com}}]{Pablo C. Lopez-Custodio}
\author[2\thanks{\tt{yuhe.gong@nottingham.ac.uk}}]{Yuhe Gong}
\author[2,3\thanks{\tt{figueredo@ieee.org}}]{Luis F.C. Figueredo}

\affil[1]{Department of Computer Science, Nottingham Trent University}
\affil[2]{School of Computer Science, University of Nottingham}
\affil[3]{Munich Institute of
Robotics and Machine Intelligence (MIRMI), Technical University of Munich (TUM)}
\affil[]{*\texttt{custodio825@gmail.com}, \textdagger\texttt{yuhe.gong@nottingham.ac.uk}, \ddag\texttt{figueredo@ieee.org}}

\begin{document}

\twocolumn[ 
  \begin{@twocolumnfalse} 

\maketitle
\begin{abstract}
Modern robotics applications require an inverse kinematics (IK) solver that is fast, robust and consistent, and that provides all possible solutions. Currently, the Franka robot arm is the most widely used manipulator in robotics research. With 7 DOFs, the IK of this robot is not only complex due to its redundancy, but also due to the link offsets at the wrist and elbow. Due to this complexity, none of the Franka IK solvers available in the literature provide satisfactory results when used in real-world applications. Therefore, in this paper we introduce GeoFIK (Geometric Franka IK), an analytical IK solver that allows the use of different joint variables to resolve the redundancy. The approach uses screw theory to describe the entire geometry of the robot, computing the Jacobian matrix prior to the joint angles. All singularities are handled. As an example of how the geometric elements obtained by the IK can be exploited, a solver with the swivel angle as the free variable is provided. Several experiments are carried out to validate the speed, robustness, and reliability of GeoFIK against three state-of-the-art solvers. The C++ code for GeoFIK is available at {\small \url{https://github.com/PabloLopezCustodio/GeoFIK}}
\end{abstract}
\vspace{0.35cm}

  \end{@twocolumnfalse} 
] 



\section{Introduction}

%
%
%

Inverse kinematics (IK) plays a pivotal role in robotics, bridging task-space goals with joint-space feasibility, execution, and performance. It is the core element of a wide range of tasks, from simple pick-and-place, grasping, and collision avoidance \cite{Chiaverini1997,Rakita2021} to highly dynamic task-and-motion planning, high-DoF planning, safety 
\cite{laha2023s,2021_Rachel_IntegratedTAMP}
and even biomechanics \cite{figueredo2021planning}.   
Unlike forward kinematics, finding IK solutions is 
challenging due to the non-linearity of the equations
and the branching of its solutions. 
While the IK of 6-DOF manipulators can have up to 16 solutions \cite{roth_ik}, the problem is more complex for redundant robots, such as 7-DOF arms, as they have infinitely many solutions organized along a continuous one-parameter manifold (self-motion) \cite{selfmotion_manifold}. 

Standard 7-DOF manipulators designed with spherical wrist and shoulder, offer trivial kinematics which allows for simple redundancy resolution. %
This is however not the case for the most widely used manipulator in robotics research of today 
\cite{2024_FrankaSurvey_Sami_RAM,Gallouedec2021PandaGym,CorkeFrankaKinematics,Oliva2022FrankaSim,2023_Elsner_TamingPanda,Feng2024TRIrobots,Lenz2023AvatarXPrize}, the Franka Research Robot. 
This manipulator features an unconventional topology with joint offsets at the wrist and elbow that break the symmetry exploited by traditional solvers. Indeed, despite the excellent work within the robotics community (e.g. tailored tools including  \cite{2023_Mario_ICRA_FrankaIdentification,Gallouedec2021PandaGym,CorkeFrankaKinematics,Oliva2022FrankaSim,2023_Elsner_TamingPanda}) a fundamental piece is missing: a fast, robust and reliable multi-parametrization analytical IK solver. 

In this context, this work presents the first IK solution for the Franka Robot based on screw theory. 
To address the challenges and the wide range of applications required to improve state-of-the-art methods and systems, 
\cite{laha2023s,2021_Rachel_IntegratedTAMP,figueredo2021planning}, our design choice aimed to satisfy the following conditions: The IK solver should   
\textbf{(i)} be \textit{faster and more robust} than any other solution, leading to solutions overlooked by other analytical methods;  
\textbf{(ii)} be \textit{more reliable}, ensuring always finding a solution if one exists;
\textbf{(iii)} return solutions even in \textit{extreme singular poses} where other approaches fail;    
\textbf{(iv)} \textit{return Jacobians} at no extra cost;  
\textbf{(v)} explore \textit{multiple free parameters} (beyond standard $q_7$), thus revealing more valid configurations than existing alternatives, including  
\textbf{(vi)} using the \textit{swivel-angle} \cite{delgado_swivel}. Obtaining the Jacobian from the IK without committing to joint angles is useful for lazy-planning strategies, dynamic and static trajectory optimization, and geometric pullbacks.

\begin{figure}[t]
    \centering
    \includegraphics[width=0.4\textwidth]{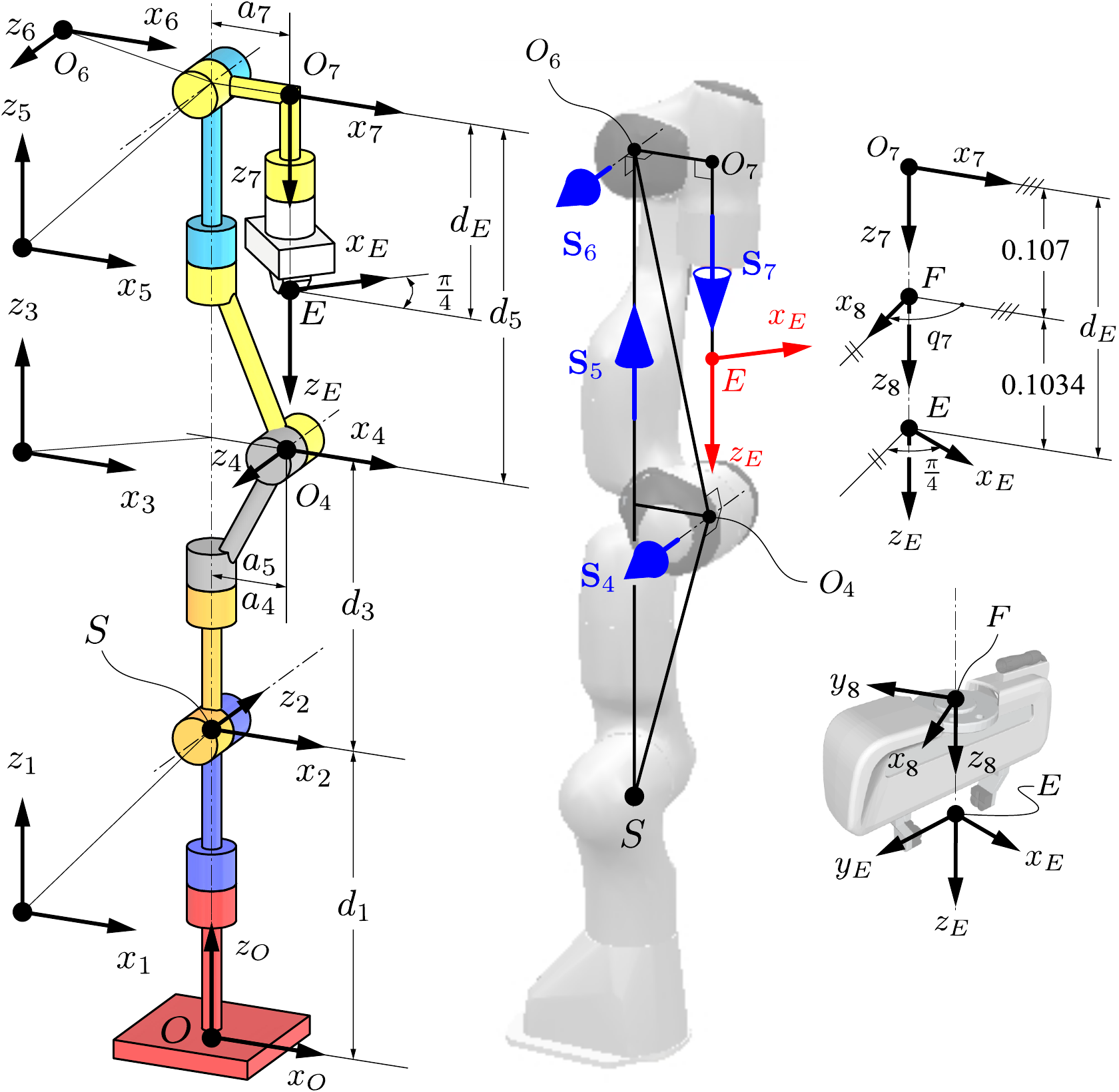}
    \caption{Home configuration of the Franka arm with $q_i=0$, $i=1,\ldots,7$, and detail of the end-effector frames. Note that both frames 8 and $E$ are rigidly attached to the gripper.}
    \label{fig:home}
\end{figure}

The Franka robot arm has firmly established itself as a reference in robotics research, with a breadth of studies showing its versatility in both industrial and academic settings. Recent surveys have shown that more than 4700 papers from leading venues such as IEEE ICRA, RA-L, T-RO, IROS, CoRL, and RSS have been using the system with a majority of the work (33~\%) focused on AI and ML.
\footnote{Manually verified papers containing keywords \textit{\scriptsize Franka robot, Emika, Panda \{arm, gripper, manipulator, cobot, robot\}, or 7 DoF Panda}, 2017--23  \cite{2024_FrankaSurvey_Sami_RAM}.} 
It is also estimated that $90$~\% and $100$~\% of the top 10 US and EU institutions in Computer Science have the system in their labs \cite{2024_FrankaSurvey_Sami_RAM}.  
The success of the Franka is mostly due to its tactile and compliance capabilities as well as increased safety capabilities, including a lower effective mass at its end-effector. See \cite{Kirschner2025_nature} for a comparison with other systems.    
As highlighted in \cite{2024_FrankaSurvey_Sami_RAM,Kirschner2025_nature}, 
achieving a safer design came with a cost, an 
offset at the wrist  and at the elbow, which makes its IK more complex than existing systems, e.g., 7-DOF KUKA arm whose wrist, elbow and shoulder are spherical. 

This complex topology (Fig.~\ref{fig:home}) reduces the effective mass, critical for safety \cite{laha2023s,Kirschner2025_nature}, but introduces additional challenges for analytical and geometric analysis. 
%
%
%
%
%
%
This challenge is reflected in the scarce number of analytical solvers for the IK of the Franka.%
\footnote{The IK can always be computed numerically by means of the {\it differential IK} using gradient-descent \cite{bruno}. However, this strategy is hundred-to-thousand-fold slower and leads to a single solution.} Indeed, up to today, only four methods were capable to derive an analytical IK, \cite{analytical_tittel,wang,liu_li,he_liu}. 
Among these, the exceptional work by He and Liu \cite{he_liu} is the only complete and functional one. It is available as a C++ application \cite{heliu_repo}. 
However, all these publications limit the free parameter to the last joint, $q_7$, which leads to an easier yet less robust analysis.    
To the best of our knowledge, no geometric solvers with other joint variables as free variable have been presented for the specific case of the Franka. Analytical solutions can be obtained from IKFast \cite{ikfast}, which precomputes algebraic expressions to automatically generate a solver. The method is general, but this comes at the cost of lacking geometric insight, being slower than geometric methods, and instability in singularity handling. The lack of geometric insight prevents the user from extracting geometric information needed in certain applications. Another option to obtain analytical solutions is to use a solver for general 6-DOF robots (e.g. \cite{pieper,ikgeo}) after one of the joints has been locked. However, this involves the definition of a new kinematic structure every time the value of the locked joint is fixed which results in additional computational time. These changes in the topology of the 6-DOF robot can also lead to unexpected behavior as we show in Sec. \ref{sec:comparison}. Finally, being able to use a 6-DOF solver for the Franka involves the development of a considerable amount of code and requires non-trivial knowledge of the robot's kinematics.


\subsection{Contributions}

Motivated by these gaps, this paper introduces 
{\bf GeoFIK} ({\bf Geo}metric {\bf F}ranka {\bf IK}), a novel solver that allows exploring free parameters (rather than only locking $q_7$) while robustly handling singularities, maximizing available solutions, and offering explicit Jacobian computation at no extra cost.
%
%

In contrast to existing methods, including \cite{he_liu,analytical_tittel,ikfast}, which find individual expressions for the joint angles, our method explores the whole geometry of the robot, finding first all the screw axes. Hence, the Jacobian matrix can be returned as a byproduct. Then, the joint angles are systematically computed using the exponential and logarithmic maps of the special orthogonal group, SO(3). Due to the use of screw theory, singularities are easily identified and handled. To our knowledge, no other IK solver for redundant manipulators works on finding the screw axes, although this is often done for mechanisms and parallel robots \cite{exechon_pc}. 

As an example of the extraction of geometric information before joint angle calculation, a fast numerical IK solver with the swivel angle as a free variable is presented. The {\it swivel angle} is roughly defined as an angle between the arm plane and a reference plane, providing a sense of posture \cite{delgado_swivel}. We demonstrate how such solver allows to preserve flexibility, guaranteeing solutions even for poses that break joint-lock-based strategies, improving robustness.  

We thoroughly compare the performance of our method against an analytical solver specifically designed for the Franka, the HeLiu solver \cite{he_liu}; a general analytical solver, IKFast \cite{ikfast}; and a 6-DOF state-of-the-art solver, IKGeo \cite{ikgeo}. HeLiu is the go-to solver for the analytical IK of the Franka, e.g. it is currently used in the MuJoCo Playground \cite{mujoco_playground}. Due to its effectiveness and generality, IKFast remains one of the most used solvers. It is, for example, the analytical solver for the popular framework MoveIt! \cite{moveit_ikfast}.
Hence, both solvers represent valuable baselines. We illustrate that GeoFIK outperforms existing methods in terms of reliability, robustness, and speed (particularly, when computing the Jacobian and joints). Furthermore, we show scenarios at singular configurations where other solvers fail. 
With real-world experiments, we also show that joint-lock-based techniques inevitably falter in discovering valid poses even for predictably easier scenarios in the middle of the robot's workspac, whereas our solver with swivel-angle as free variable consistently finds solutions and remains computationally efficient.  
%

%

This paper is organized as follows: Sec. \ref{sec:problem} introduces the problem statement and general method; Secs. \ref{sec:sol_q7} to \ref{sec:sol_swivel} present the solutions locking $q_7$, $q_6$, $q_4$ and swivel angle, respectively; Sec. \ref{sec:sing} discusses singularity handling; Sec. \ref{sec:comparison} provides comparisons with baselines; Sec. \ref{sec:experiments} shows a real-world experiment; and conclusions are drawn in Sec. \ref{sec:conclusions}.

\section{Problem Formulation and Method Description} \label{sec:problem}


This section defines our IK problem, and outlines our screw-theory-based approach.%
\footnote{%
Fig. \ref{fig:home} shows the frames and DH parameters of the Franka arm. We use the frame $E$ as end-effector frame. When the gripper is connected, $E$ matches the frame \texttt{panda\textunderscore hand\textunderscore tcp} in Franka Description \cite{fci_description}. 
We rename the centre of the shoulder as $S$. 
}  

\newcommand{\realset}{\mathbb{R}}
\newcommand{\SO}[1]{\mathrm{SO}(#1)}
\newcommand{\SE}[1]{\mathrm{SE}(#1)}
\newcommand{\jointsset}{\mathbb{S}}
\newcommand{\screw}[2]{{}^{#1}\mathbf{S}_{#2}}
\newcommand{\screwdir}[2]{{}^{#1}\uvec{s}_{#2}}
\newcommand{\screwpt}[2]{{}^{#1}\mathbf{r}_{#2}}
\newcommand{\myvector}[1]{\mathbf{#1}}
\newcommand{\mymatrix}[1]{\mathbf{#1}}

\noindent \textbf{Problem Definition}:  
Given a desired pose of the end-effector frame%
\footnote{Notation: $\pv{O}{r}{P/Q}\in\mathbb{R}^3$ is the position vector of point $P$ with respect to point $Q$ measured in frame $O$. $\rotmat{O}{E}\in\mathrm{SO}(3)$ is the rotation matrix defining the orientation of frame $E$ with respect to frame $O$. Unit vectors wear a hat, $\hat{\mathbf{u}}\in\mathcal{S}^2$. We use the parallel symbol ``$\|$" to avoid normalising the right-hand side when the left-hand side is a unit vector, e.i. $\uvec{u}\,\|\,\mathbf{r}$ means $\uvec{u}=\mathbf{r}/|\mathbf{r}|$.} 
 $E$ w.r.t. the base frame $O$ defined by
${}^O\mathbf{r}_{E/O} {\in} \realset^3$  
and 
$\rotmat{O}{E} {\in}  \SO{3}$ 
the goal is to find a set of joint-angles 
$q_i {\in} \jointsset$, with $i{=}1,{\ldots},7$ that leads to the desired pose of frame $E$.    

Given the 1-DOF redundancy, one additional constraint must be provided. We provide solutions that use $q_4,q_6,q_7$, or the swivel angle to resolve this redundancy. No $q_5$ solver is provided since it is geometrically the most complex and will be researched in future work.

\noindent\textbf{Screw Theory-based IK Solution:} Unlike traditional IK solvers that rely on purely algebraic formulations, our approach uses screw theory, offering a geometric perspective on the solution. The method is based on determining the screw axes of the manipulator
$\screw{O}{i} = (  \screwdir{O}{i}; ~ \screwpt{O}{i} \times \screwdir{O}{i} ), \ \  i=1,\ldots,7$,
where $\screwdir{O}{i}$ is the direction of the axis, 
and $\screwpt{O}{i}$ is the position of {\it any} point along the axis. 

By using this formulation, the \textbf{Jacobian} can be computed even without prior knowledge of joint angles as 
$\mymatrix{J}=\mathrm{Adj}(\mathbf{I}_3,-\pv{O}{r}{E/O})\big[{}^O\mathbf{S}_1,\ldots,{}^O\mathbf{S}_7\big]$, 
where $\mathrm{Adj}(\cdot,\cdot)$ is the adjoint representation of an element of $\SE{3}$ defined by a pair in $\mathrm{SO}(3)\times\mathbb{R}^3$, 
see \cite[Sec.~4.2]{selig}.

We first exploit the spherical shoulder joint structure of the Franka, and hence we first find screw-axes $\screw{O}{4}$, $\screw{O}{5}$, and $\screw{O}{6}$. The axis of the last joint, $\screw{O}{7}$, is fully determined by the input, i.e., 
$\screwdir{O}{7} = {}^O\hatk_E$ 
and 
$\screwpt{O}{7}={}^O\mathrm{r}_{E/O}$.

Once those are known, the spherical shoulder is solved as
\begin{gather*}
    {}^O\uvec{s}_3\,\|\,\exp\big(\arctan(a_4/d_3){}^O\uvec{s}_4\big){}^O\mathbf{r}_{O_4/S},\;{}^O\uvec{s}_1=(0,0,1)^{\top},
    \\
    {}^O\uvec{s}_2^{(i)}\,\|\,(-1)^{i}\big({}^O\uvec{s}_1\times{}^O\uvec{s}_3\big),\;i=1,2,
    \\
    {}^O\mathbf{r}_{j}=\pv{O}{r}{S/O}=(0,0,d_1)^{\top},\;j=1,2,3,
\end{gather*}

Once the screw axes are known, we compute the {\bf joint angles} by comparing how these axes rotate away from a known home configuration ($q_i{=}0$).  
Let $\screwdir{O}{i,0}$ be the directions of the axes at home configuration, then, from  Fig.~\ref{fig:home}:   
\[
\big[{}^O\uvec{s}_{1,0},\ldots,{}^O\uvec{s}_{7,0}\big]=\left[\begin{array}{ccccccc}
     0 & 0 & 0 & 0 & 0 & 0 & 0  \\
     0 & 1 & 0 & -1 & 0 & -1 & 0 \\
     1 & 0 & 1 & 0 & 1 & 0 & -1
\end{array}\right]
\]
With this information, we can compute the exponential map in $\SO{3}$ to align each axis step by step. This process is shown in Algorithm \ref{alg:joint_angles}.  Note that line 4 is a simplification of $q_j\gets\big|\log\big(\rotmat{O}{j+1}^{\top}\rotmat{O}{j}\big)\big|$.


\begin{algorithm}[t]
    \DontPrintSemicolon
    \KwIn{${}^O\uvec{s}_{i,0}$, ${}^O\uvec{s}_{i}$, $i=1,\ldots,7,\,{}^O\hati_E$}
    \KwOut{$q_1,\ldots,q_7$}
    $\uvec{u}_i\gets{}^O\uvec{s}_{i,0}$, $i=1,\ldots,7$,\,\,$\uvec{u}_8\gets(\sqrt{2}/2,\sqrt{2}/2,0)^{\top}$\;
    $\uvec{g}_i\gets{}^O\uvec{s}_i$, $i=1,\ldots,7$,\,\,$\uvec{g}_8\gets{}^O\hati_E$\;
    \For{$j \gets 1$ \textnormal{to} $7$}{
    $q_j \gets \mathrm{atan2}\big((\uvec{u}_{j+1}\times\uvec{g}_{j+1})\cdot\uvec{u}_j,\, \uvec{u}_{j+1}\cdot\uvec{g}_{j+1}\big)$\;
    $\big[\uvec{u}_{j+1},\ldots\uvec{u}_8\big] \gets \exp\big(q_j\uvec{u}_j\big)\big[\uvec{u}_{j+1},\ldots\uvec{u}_8\big]$\;
    }
  \caption{Calculating the joint angles $q_i$ from the directions of the joint axes ${}^O\uvec{s}_i$, $i=1,\ldots,7$}\label{alg:joint_angles}
\end{algorithm}

In the Secs. \ref{sec:sol_q7}, \ref{sec:sol_q6}, \ref{sec:sol_q4} and \ref{sec:sol_swivel} we present solvers that have, respectively, $q_7$, $q_6$, $q_4$ and swivel angle as free variable. Each solver has as input the value of the free variable, and ${}^O\hati_E$, ${}^O\uvec{s}_7={}^O\hatk_E$, and ${}^O\mathbf{r}_{O_7/O}={}^O\mathbf{r}_{E/O}-d_E{}^O\hatk_E$ (and ${}^O\mathbf{r}_{O_7/S}$). The goal is to find ${}^O\mathbf{S}_4$, ${}^O\mathbf{S}_5$ and ${}^O\mathbf{S}_6$.


\section{Solution locking joint 7}\label{sec:sol_q7}
It can be seen that ${}^O\uvec{s}_6=\exp\big((\frac{3}{4}\pi-q_7\big){}^O\uvec{s}_7){}^{O}\hati_E$ and ${}^O\mathbf{r}_6=\pv{O}{r}{O_6/O}={}^O\mathbf{r}_{O_7/O}-a_7{}^O\uvec{s}_6\times{}^O\uvec{s}_7$.
\begin{figure}[b]%
    \centering%
    \includegraphics[width=0.3\textwidth]{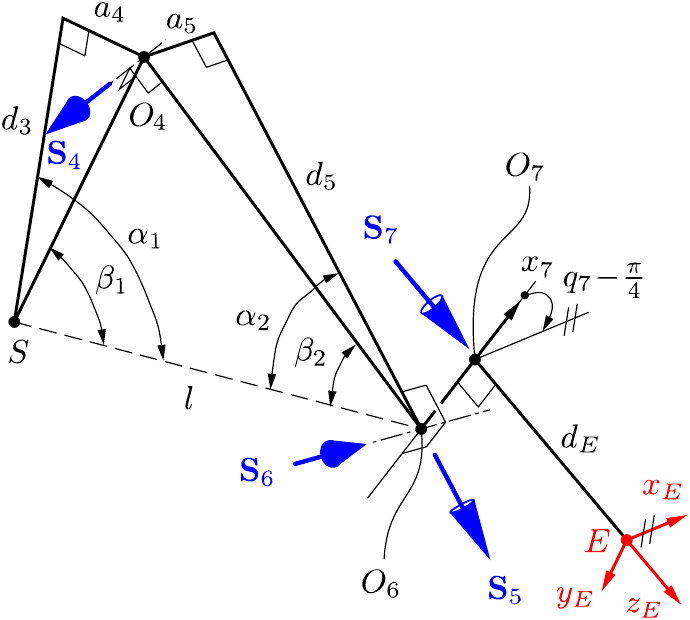}
    \caption{Franka Robot Geometry for the IK solution fixing $q_7$.}
    \label{fig:q7}
\end{figure}

Consider $\pv{O}{r}{O_6/S}=\pv{O}{r}{O_6/S}-\pv{O}{r}{S/O}$ and let $l:=|\pv{O}{r}{O_6/S}|$, $b_1:=\sqrt{d_3^2+a_4^2}$, $b_2:=\sqrt{d_5^2+a_5^2}$, $\rho_1:=\arctan(a_4/d_3)$ and $\rho_2:=\arctan(a_5/d_5)$ in Fig. \ref{fig:q7}, then:
\begin{gather}\label{eq:alpha_2}
    \alpha_2^{(1)}=\rho_2+\beta_2,\;\;\alpha_2^{(2)}=\rho_2-\beta_2,
    \\
    \text{with}\;\;\beta_2:=\arccos\left((b_1^2-b_2^2-l^2)/(-2b_2l)\right). \nonumber
\end{gather}  

These two solutions correspond to the elbow-up and elbow-down assemblies of triangle $SO_4O_6$. However, as noted in \cite{he_liu}, due to link interference, solution 2 only happens when $SO_4O_6$ is almost flat and $d_3+d_5<l<b_1+b_2$, Fig. \ref{fig:sing2} shows one of such scenarios.

Define a frame $C$ with ${}^O\hat{\mathbf{k}}_C\|{}^O\mathbf{r}_{S/O_6}$ and ${}^O\hati_C\|{}^O\hat{\mathbf{k}}_C\times{}^O\hat{\mathbf{s}}_6$. This allows us to easily write ${}^C\uvec{s}_5=-(\sin\alpha_2\cos\gamma,$ $\sin\alpha_2\sin\gamma,$ $\cos\alpha_2)^{\top}$, where, if ${}^C\uvec{s}_6={}^C_O\mathbf{R}{}^O\uvec{s}_6=(0,s_{6,y},s_{6,z})$, imposing the constraint ${}^C\uvec{s}_5{\cdot{}}^C\uvec{s}_6{=}0$ results in
\[
\gamma^{(1)}=\arcsin\left(A_{\gamma}\right),\, \gamma^{(2)}=\pi-\arcsin\left(A_{\gamma}\right),
\]
where, $A_{\gamma}:=-(s_{6,z}/s_{6,y})\cot\alpha_2$. If $A_{\gamma}>1$, the kinematic chain cannot be assembled, and if $A_{\gamma}=1$ the two solutions for $\gamma$ are repeated. Finally, $^{O}\uvec{s}_5={}^O_{C}\mathbf{R}{}^C\uvec{s}_5$ and ${}^O\mathbf{r}_5={}^O\mathbf{r}_6$. The remaining parameters are easily found as follows:
\[
{}^O\uvec{s}_4\,\|\,{}^O\uvec{s}_5\times{}^{O}\mathbf{r}_6,\; {}^O\mathbf{r}_4={}^O\mathbf{r}_6-d_5{}^O\uvec{s}_5 + a_5{}^O\uvec{s}_5\times{}^O\uvec{s}_4.
\]
Since $\alpha_2$, $\gamma$ and $\uv{O}{s}{2}$ can have up to two solutions each, the maximum number of solutions is $2^3=8$.


\section{Solution locking joint 6}\label{sec:sol_q6}

\begin{figure}[t]
    \centering
    \includegraphics[width=0.35\textwidth]{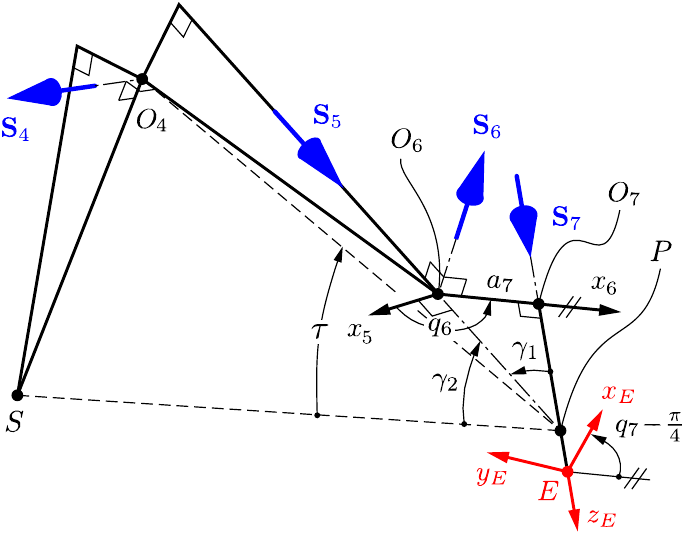}
    \caption{Geometry of the Franka for the IK fixing $q_6 {\notin} \{0,\pi\}$.}
    \label{fig:q6}
\end{figure}

Define $\gamma_1:=\pi-q_6$ as the angle between $\mathbf{S}_5$ and $\mathbf{S}_7$, and let $P$ be their intersection point as shown in Fig. \ref{fig:q6}. The following two cases are analysed separately. 

\begin{figure}[t]
    \centering
    \includegraphics[width=0.4\textwidth]{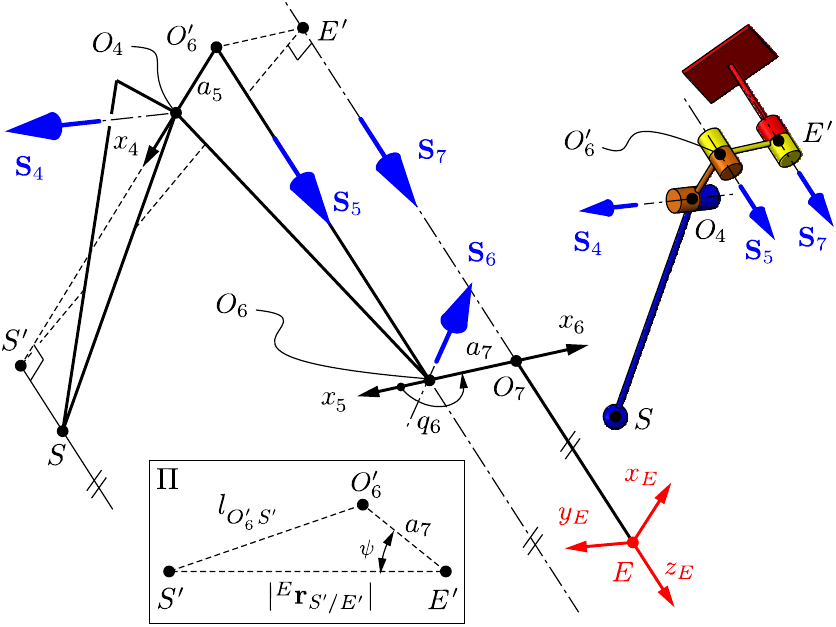}
    \caption{Geometry of the Franka for the IK fixing $q_6 {\in}\{0,\pi\}$.}
    \label{fig:q6_parallel}
\end{figure}

{\bf $\mathbf{S}_5$ and $\mathbf{S}_7$ are not parallel:} It can be seen that ${}^O\mathbf{r}_{P/S}=\pv{O}{r}{O_7/S}+(a_7/\tan\gamma_1)\uk{O}{E}$. Let $\gamma_2$ be the angle from $\mathbf{r}_{S/P}$ to $-\mathbf{S}_5$ measured about $-\uvec{s}_4$ as shown in Fig. \ref{fig:q6}. The goal is to first find $\gamma_2$ using the polygon $SO_4O_6P$ and then use the result to find $\uvec{s}_5$.

Let $l_C$ be the projection of $\mathbf{r}_{P/O_6}$ on $\mathbf{S}_5$, so $l_C=a_7/\sin\gamma_1$. The angle $\tau$, defined as $\angle SPO_4$ can be easily found by solving first the triangle $PO_6O_4$ followed by triangle $SPO_4$. It can be seen that 
\[ \gamma_2^{(1)}=\begin{cases} 
      \tau+\arctan(a_5/(d_5+l_C))+\pi & \text{if}\;\;l_C<-d_5, \\
      \tau+\arctan(a_5/(d_5+l_C)) & \text{otherwise} 
   \end{cases}
\]
The extreme case when $SPO_4$ is almost flat, i.e. $d_3 + d_5 + l_C < |\pv{O}{r}{S/P}| $$< b_1 + (a_5^2+(l_C+d_5)^2)^{1/2}$, provides a second solution $\gamma_2^{(2)}$ in which $\tau$ appears with negative sign. 

Now we can write an equation to constrain the angle between $\mathbf{r}_{S/P}$ and $\mathbf{S}_5$ to $\gamma_2$. In frame $E$ it is easy to write $\uv{E}{s}{5}=(-\sin\gamma_1\cos u, -\sin\gamma_1\sin u, \cos\gamma_1)^{\top}$, where $u:=5\pi/4-q_7$. Then, the following equation in $u$ is to be solved:
\[
-\uv{E}{s}{5}\cdot\pv{E}{r}{S/P}=\uv{E}{s}{5}\cdot\big({}^{O}_E\mathbf{R}\big)^{\top}\pv{O}{r}{P/S}=|\pv{O}{r}{P/S}|\cos\gamma_2
\]
which has the two solutions $u^{(1)}=u_1-u_2$ and $u^{(2)}=\pi-u_1-u_2$, where, if $\pv{E}{r}{S/P}=(C_1,C_2,C_3)^{\top}$:
\[
\begin{split}
    & u_1 := \arcsin\left(\frac{|\pv{O}{r}{P/S}|\cos\gamma_2+C_3\cos\gamma_1}{|\sin\gamma_1|\sqrt{C_1^2+C_2^2}}\right),\,
    \\
    &u_2:=\mathrm{atan2}\big(C_1\sin\gamma_1, C_2\sin\gamma_1\big)
\end{split}
\]

With $\pv{O}{r}{P/O}=\pv{O}{r}{P/S}+\pv{O}{r}{S/O}$, the remaining parameters are found easily:
\[
\begin{split}
    & \pv{O}{r}{5}=\pv{O}{r}{6}=\pv{O}{r}{O_6/O}=\pv{O}{r}{P/O}-l_C\uv{O}{s}{5},\;
    \\
    &\uv{O}{s}{6}\,\|\,\uv{O}{k}{E}\times(\pv{O}{r}{O_7/O}-\pv{O}{r}{O_6/O}),\;\uv{O}{s}{4}\,\|\,\uv{O}{s}{5}\times\pv{O}{r}{P/S},
    \\
    &\pv{O}{r}{4}=\pv{O}{r}{O_4/O}=\pv{O}{r}{P/O}-(d_5+l_C)\uv{O}{s}{5}+a_5\uv{O}{s}{5}\times\uv{O}{s}{4}
\end{split}
\]
Since $\gamma_2$, $u$ and $\uv{O}{s}{2}$ have up to two solutions, the maximum number of solutions is $(2)^3=8$.

{\bf $\mathbf{S}_5$ and $\mathbf{S}_7$ are parallel:} Since $q_6\in\{0,\pi\}$, we define $\sigma=\cos q_6\in\{1,-1\}$. Consider the plane $\Pi$ which is perpendicular to $\mathbf{S}_5$ and $\mathbf{S}_7$ and contains $\mathbf{S}_4$, see Fig. \ref{fig:q6_parallel}. Let $E'$, $O_6'$ and $S'$ be the respective projections of $E$, $O_6$ and $S$ onto $\Pi$. Note that $\pv{E}{r}{E'/E}=(0,0,-d_E+\sigma d_5)^{\top}$. Since $E'$ is known, the problem is reduced to solving the equivalent kinematic chain shown in Fig. \ref{fig:q6_parallel} by working on plane $\Pi$ and in frame $E$. 

\begin{figure}[t]
    \centering
    \includegraphics[width=0.43\textwidth]{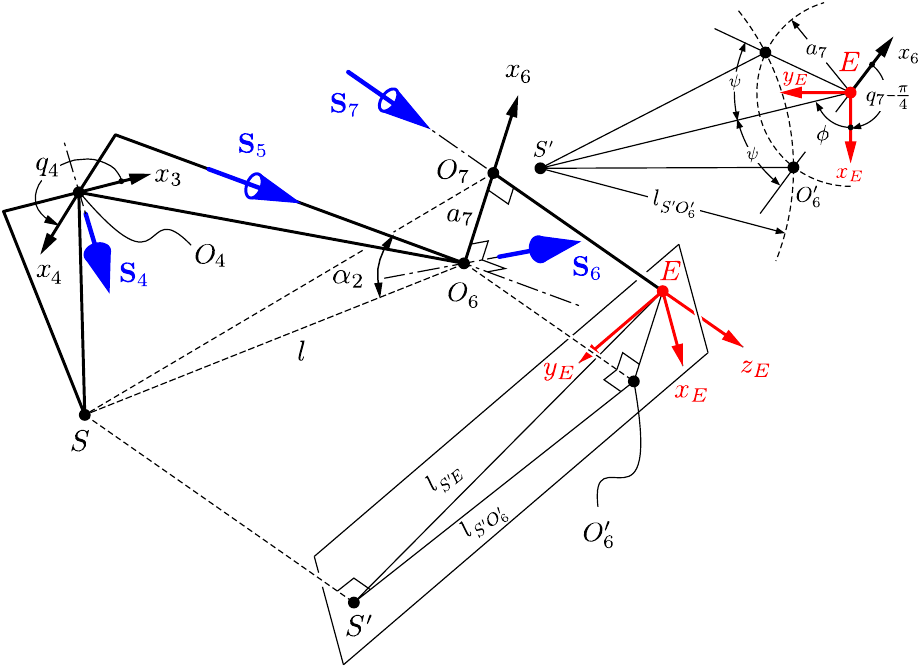}
    \caption{Geometry of the Franka for the IK solution fixing $q_4$.}
    \label{fig:q4}
\end{figure}

It can be seen that $\pv{E}{r}{S'/E'}=\pv{E}{r}{S/E'}-(\pv{E}{r}{S/E'}\cdot\uv{E}{k}{E})\uv{E}{k}{E}$, where $\pv{E}{r}{S/E'}=\pv{E}{r}{S/E}-\pv{E}{r}{E'/E}$. Due to the two possible ways of reaching $S$ from $\Pi$, the distance from $O_6'$ to $S'$ along $x_4$ is $l_{O_6'S'}^{(i)}=a_5+(-1)^{i}\big(d_3^2+a_4^2-\big(\pv{E}{r}{S/E'}\cdot\uv{E}{k}{E}\big)^2\big)^{1/2}$, $i=1,2$. This leads to two solutions for $\psi$, the angle $\angle S'E'O_6'$:
\[
\psi^{(i)}=\arccos\left[\frac{(l_{O_6'S'}^{(i)})^2-a_7^2-|\pv{E}{r}{S'/E'}|^2}{-2a_7|\pv{E}{r}{S'/E'}|}\right],\;i=1,2.
\]
Similarly, due to the two different ways of assembling the triangle $S'E'O_6'$, two solutions for the position of $O_6'$ appear for each solution of $\psi$:
\[
    \pv{E}{r}{O_6'/E'}^{(i)}=a_7\exp\big((-1)^{i}\psi\uv{E}{k}{E}\big)\pv{E}{r}{S'/E'}/|\pv{E}{r}{S'/E'}|,\; i=1,2.
\]
It follows that $\pv{E}{r}{6}=\pv{E}{r}{E'/E}+\pv{E}{r}{O_6'/E'}-\sigma d_5\uv{E}{k}{E}$ and $\uv{E}{s}{6}\,\|\,\pv{E}{r}{O_6'/E'}\times\uv{E}{k}{E}$. For $\pv{E}{S}{5}$, it can be seen that $\uv{E}{s}{5}=-\sigma\uk{E}{E}$ and $\pv{E}{r}{5}=\pv{E}{r}{6}$. To find ${}^E\mathbf{S}_4$, consider $\ui{E}{4}\,\|\,\mathrm{sign}(l_{O_6'S'})\big(\pv{E}{r}{S'/E'}-\pv{E}{r}{O_6'/E'}\big)$, then $\pv{E}{r}{4}=\pv{E}{r}{E'/E}+\pv{E}{r}{O_6'/E'}+a_5\ui{E}{4}$, and $\uv{E}{s}{4}=\ui{E}{4}\times\uv{E}{s}{5}$. All the screw axes are then transformed back to frame $O$.

Since $\psi$, $\pv{E}{r}{O_6'/E'}$ and $\uv{O}{s}{2}$ can have up to two solutions each, the maximum number of solutions is $2^3=8$.


\begin{figure}[t]
    \centering
    \includegraphics[width=0.35\textwidth]{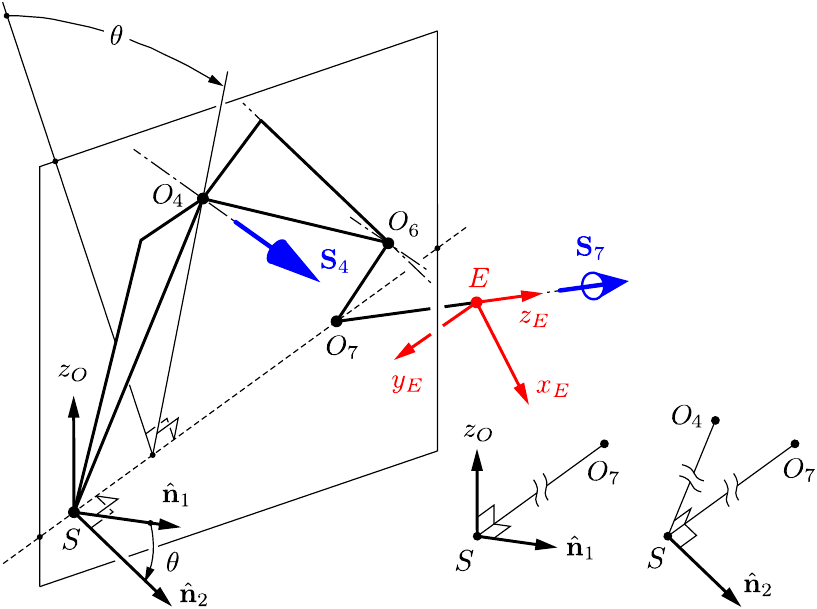}
    \caption{Geometry of the Franka arm for the IK solution fixing the swivel angle $\theta$.}
    \label{fig:swivel}
\end{figure}

\section{Solution locking joint 4}\label{sec:sol_q4}

Since in this case $q_4$ is given, the triangle $SO_4O_6$ can be easily solved to obtain $l:=|\mathbf{r}_{S/O_6}|$ and $\alpha_2$, see Fig. \ref{fig:q4}. We can then locate $O_6$ on a sphere of radius $l$ and centre at $S$. From the pose of frame $E$, we can also constrain $O_6$ to a known circle of radius $a_7$ and centre at $O_7$. Hence, the problem is solved by finding the intersection of this circle with the sphere. This is reduced to the intersection of two circles on the $X_EY_E-$plane.


As shown in Fig. \ref{fig:q4}, let $S'$ and $O_6'$ be the respective projections of points $S$ and $O_6$ onto the $X_EY_E-$plane, then define $l_{S'O_6'}:=|\mathbf{r}_{S'/O_6'}|=(l^2-(\pv{E}{r}{O_7/S}\cdot\uk{E}{E})^2)^{1/2}$ and $l_{S'E}:=|\mathbf{r}_{S'/E}|=((\pv{E}{r}{O_7/S}\cdot\ui{E}{E})^2+(\pv{E}{r}{O_7/S}\cdot\uj{E}{E})^2)^{1/2}$. 

To find $O_6'$, consider $\phi$, the angle from $x_E$ to $\mathbf{r}_{S'/E}$ about $z_E$, and $\psi$, the angle $\angle S'EO_6'$, as shown in Fig. \ref{fig:q4}. It can be seen that $\phi=\mathrm{atan2}(\pv{E}{r}{S/O_7}\cdot\uj{E}{E},\pv{E}{r}{S/O_7}\cdot\ui{E}{E})$ and $\psi=\arccos\big((l_{S'E}^2+a_7^2-l_{S'O_6'}^2)/(2a_7l_{S'E})\big)$, leading to the two solutions $q_7^{(i)}=(-1)^i\psi-\phi-3\pi/4$, $i=1,2$.

With $O'_6$ in hand, it is easy to find ${}^O\mathbf{S}_6$ with $\uv{O}{s}{6}=\rotmat{O}{E}\exp\big((-q_7+3\pi/4)\uk{E}{E}\big)\ui{E}{E}$ and $\pv{O}{r}{O_6/O} =$ $\rotmat{O}{E}\pv{E}{r}{O_6/O_7}+\pv{O}{r}{O_7/O}$, where $\pv{E}{r}{O_6/O_7}=-a_7\exp\big((-q_7+\pi/4)\uk{E}{E}\big)\ui{E}{E}$.

Since $q_7$, $\pv{O}{S}{6}$ and $\alpha_2$ are known now, we can proceed as we did in Sec. \ref{sec:sol_q7}, where $q_7$ was a free parameter. Following the same procedure, we find two solutions of $\uv{O}{s}{5}$ per each value of $q_7$. Since $q_7$, $\uv{O}{s}{5}$ and $\uv{O}{s}{2}$ can have up to two solutions each, the total number of solutions is $2^3=8$.


\section{Solution for swivel angle}\label{sec:sol_swivel}

Due to the lack of a spherical wrist, the swivel angle $\theta$ can be defined using either $O_6$ or $O_7$ as ``wrist centre". Let this be $O_7$, like in \cite{yang_swivel}. Then the swivel angle is defined between the plane containing $z_O$ and $O_7$, and the plane containing $O_4$, $S$ and $O_7$ as shown in Fig. \ref{fig:swivel}. Hence, if:
\begin{gather}\label{eq:swivel_angle_def}
\uv{O}{n}{1}\,\|\,\pv{O}{r}{O_7/S}\times\uk{O}{O}\;\;\text{and}\;\;\uv{O}{n}{2}\,\|\,\sigma\pv{O}{r}{O_7/S}\times\pv{O}{r}{O_4/S}
\\
\mathrm{with}\;\;\sigma= \mathrm{sign}\big(\pv{O}{r}{O_7/S}\times\pv{O}{r}{O_4/S}\cdot\uv{O}{s}{4}\big)
\end{gather}
then $\theta$ is the angle from $\uv{O}{n}{1}$ to $\uv{O}{n}{2}$ measured about $\pv{O}{r}{O_7/S}$. 

In order to find all the solutions that have the desired value of $\theta$, we discretise the range of $q_7$, calculate $\theta$ for each point, pick the ones that are close enough to the desired value, and finally we interpolate to find the optimal values of $q_7$. $q_7$ is used as free variable since, by doing so, the first singularity discussed in Sec. \ref{sec:sing} is avoided.


We do not need to solve the entire IK with $q_7$ as input, we only obtain the strictly necessary geometric elements that allow us to calculate $\theta$. That means proceeding as done in Sec. \ref{sec:sol_q7} until $\pv{O}{r}{7}$, $\pv{O}{r}{6}$ and $\uv{O}{s}{4}$ are found. To speed up the process, we ignore the elbow-down solution with $\alpha_2^{(2)}$ in Eq. (\ref{eq:alpha_2}), which appears in the unlikely cases where $d_3+d_5<l<b_1+b_2$.


\begin{figure}[t]
    \centering
    \includegraphics[width=0.35\textwidth]{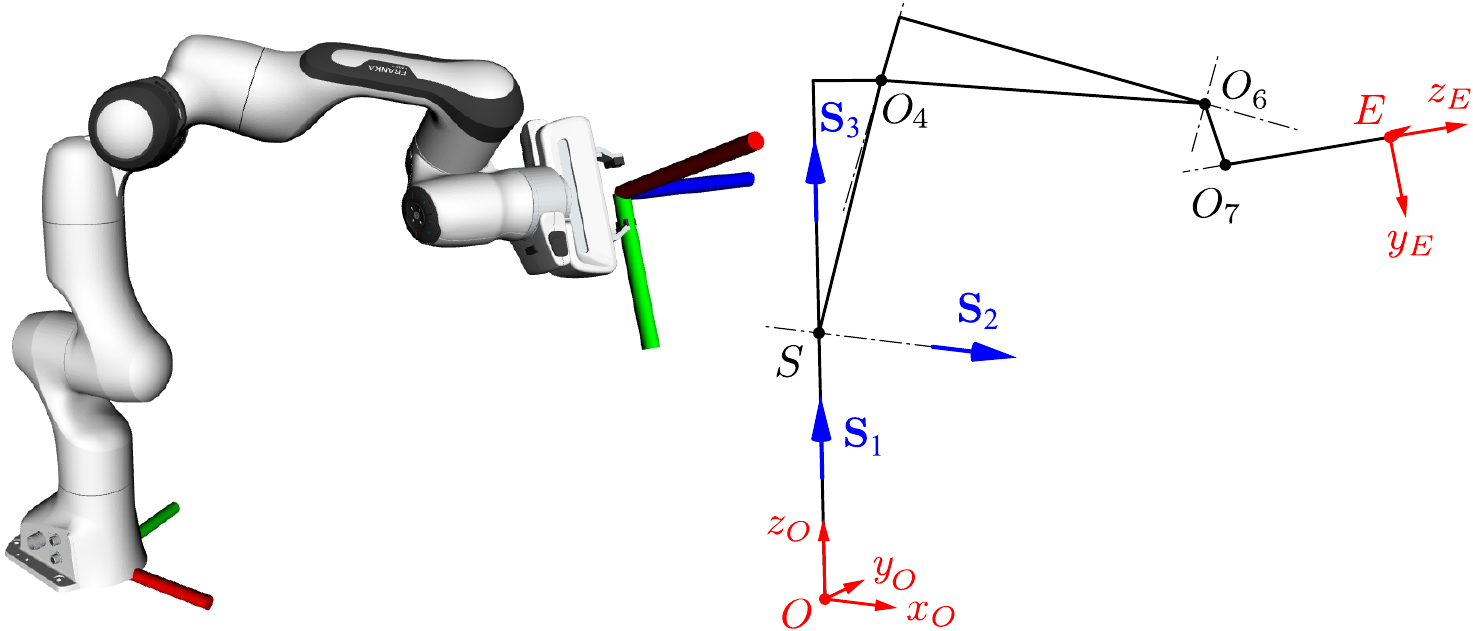}
    \caption{An example of flat shoulder singularity.}
    \label{fig:sing1}
\end{figure}
\begin{figure}[t]
    \centering
    \includegraphics[width=0.48\textwidth]{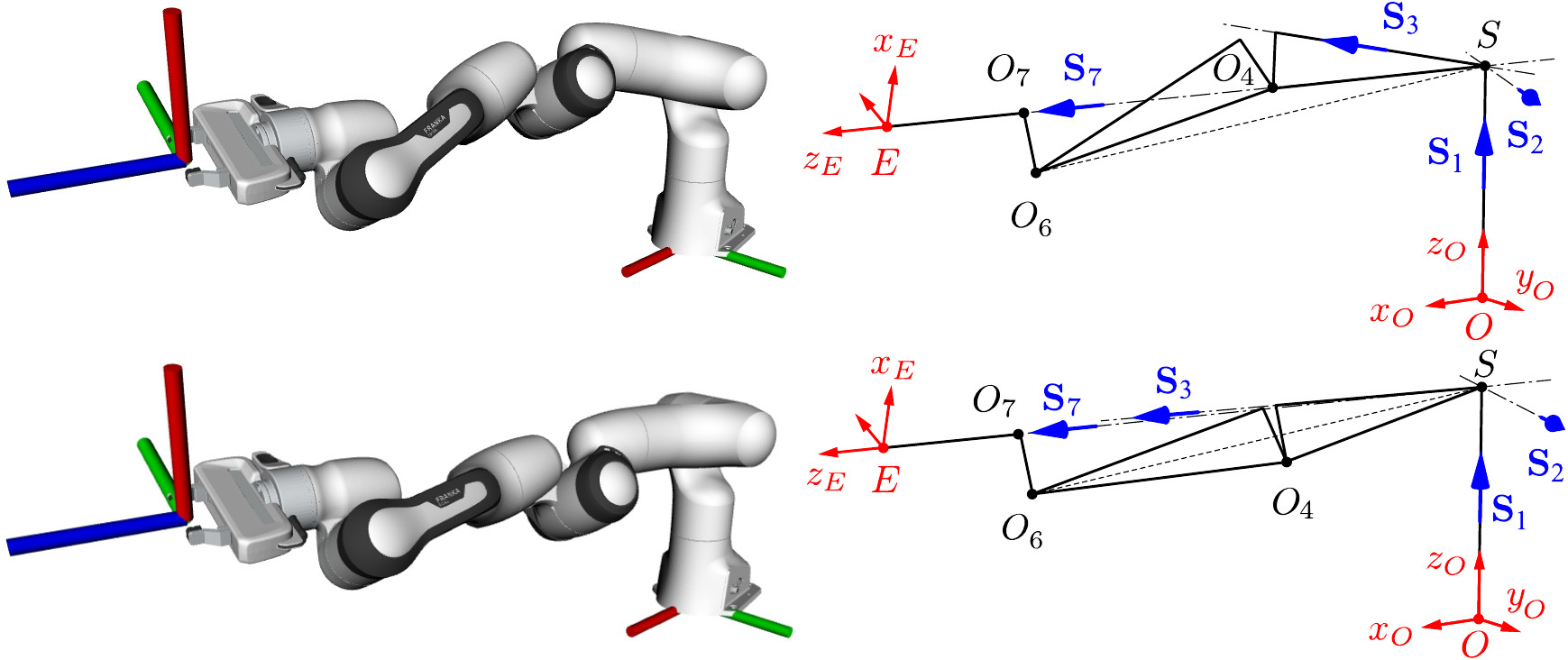}
    \caption{Elbow-up and elbow-down solutions in a Shoulder-on-$\mathbf{S}_7$ singularity. Note that these are {\it not} work-space boundary singularities as the arm is not fully stretched.}
    \label{fig:sing2}
\end{figure}

\section{Singularity analysis}\label{sec:sing}

Singularities in 7-DOF robots can be classified as kinematic and algorithmic singularities \cite{gosselin_abb_yumi}. {\it Kinematic singularities} are configurations in which $\mathrm{rank}(\mathbf{J}(\mathbf{q}))<6$. This rank-deficiency are the result of special geometric conditions between the screw axes. For our IK solvers, these conditions will appear as extreme cases of triangles becoming flat or axes becoming parallel. To provide numerical stability in those cases, GeoFIK includes a constant {\small \texttt{SING\_TOL}} which defines how close to the extreme value a variable can be to still attempt a solution, e.g. if $x>1$ but $x-1<${\small \texttt{SING\_TOL}}, $\arccos{x}=0$ is considered. An example occurs when $\mathbf{S}_4$ and $\mathbf{S}_6$ are parallel and coplanar with $S$. As a consequence, $\mathbf{S}_5$, $\mathbf{S}_6$ and $S$ are coplanar too, resulting in $\mathrm{rank}(\mathbf{J}(\mathbf{q}))=5$ and the triangle $SO_4O_6$ becoming flat. It can be seen that work-space boundary configurations are kinematic singularities. In general, since the Franka has 7 joints, two special geometric conditions are needed to take $\mathrm{rank}(\mathbf{J})$ below 6.

In {\it algorithmic singularities} the self-motion no longer affects the variable we want to fix to resolve the redundancy. If $q_i$ is being fixed, $\mathrm{rank}(\mathbf{J}(\mathbf{q}))=6$ can hold, but if $\mathbf{S}_i$ is linearly independent from the other screw axes, then it is not possible to have $q_i$ as free variable. In the Franka arm with the redundant variables considered in this paper, such cases can only occur in the following scenario:

\noindent{\bf Shoulder-on-$\mathbf{S}_7$ singularity}: If $\mathbf{S}_7$ intersects the spherical shoulder at $S$, the robot experiences self-motion with the polygon $SO_4O_6O_7$ spinning around $\mathbf{S}_7$ without affecting $q_4$, $q_5$ and $q_6$. Hence, only $q_7$ can be used as free variable with the resulting 6-DOF robot being in regular configuration. If GeoFIK detects that $S\in\mathbf{S}_7$, the solver with $q_7$ as free variable is called with an emergency value of $q_7$ provided by the user. Fig. \ref{fig:sing2} shows an example of this singularity.

Theoretically, other algorithmic singularities can appear whenever a pair of joint axes become collinear or a joint axis intersects $S$. However, due to the link-lengths and joint limits of the Franka, this will only occur in the two scenarios described above. As noted in \cite{CorkeFrankaKinematics}, the home configuration (Fig. \ref{fig:home}) is a kinematic singularity with $\mathrm{rank}(\mathbf{J}(\mathbf{q}))=5$, but it is also algorithmic-singular as $\mathbf{S}_5$ intersects $S$. This case is not implemented in GeoFIK since it violates joint limits.

As reported in \cite{he_liu,analytical_tittel}, if $q_2=0$ the self-motion is reduced to $\mathbf{S}_2$ spinning about $\mathbf{S}_1=\mathbf{S}_3$ (Fig. \ref{fig:sing1}). However, this configuration is not a kinematic singularity since, in general, $\mathrm{rank}(\mathbf{J})=6$. The robot is neither in an algorithmic singularity as we ignore the first 3 joints to obtain the rest of the joint axes. However, the shoulder is in a singularity since $\mathrm{rank}([\mathbf{S}_1,\mathbf{S}_2,\mathbf{S}_3])=2$. Hence, we refer to this configuration as {\bf flat shoulder singularity}. Only the definition of $\mathbf{S}_2$ is affected, with $q_1$ becoming a free choice. The user can provide an emergency value of $q_1$ to be used in this situation.



Another kind of singularity occurs in the swivel angle definition (\ref{eq:swivel_angle_def}). Clearly, $\theta$ is not defined if $\mathbf{r}_{O_7/S}$ is parallel to $z_O$. This is the unavoidable singularity of any swivel angle definition \cite{sew}. GeoFIK prints an error message in those cases. Note that the sign $\sigma$ prevents the sudden change of direction of the plane containing $S$, $O_7$ and $O_4$. In addition, if these three points are collinear, $\uvec{n}_2=\uvec{s}_4$ is chosen.

\begin{table}[t!]
\centering
\caption{Solver Comparison for  Singular Configurations (angles in degrees). Solutions in orange indicate larger errors (EE position) ${\geq} 1$~cm, and in red,  violations of joint limits.}\label{tab:sing}
\begin{scriptsize}
\begin{tabular}{p{0.3cm}c}
    \noalign{\hrule width 8.5cm height 1.5pt}
    \multicolumn{2}{c}{\rule{0pt}{8pt} \textbf{Case 1 (Flat shoulder, solvers with q7 as free variable)}} \\ 
    \noalign{\hrule width 8.5cm}
    {} & 
        $\begin{array}{p{\wcolzer}p{\wcolone}p{\wcoltwo}p{\wcolthr}p{\wcolfou}p{\wcolfiv}p{\wcolsix}p{\wcolsev}}
        sol & \ensuremath{q_1} & \ensuremath{q_2}  & \ensuremath{q_3} & \ensuremath{q_4} & \ensuremath{q_5} & \ensuremath{q_6} & \ensuremath{q_7} 
        \end{array}  $
    \\
    \noalign{\hrule width 8.5cm}
    GeoFIK (ours) &
        $\begin{array}{p{\wcolzer}p{\wcolone}p{\wcoltwo}p{\wcolthr}p{\wcolfou}p{\wcolfiv}p{\wcolsix}p{\wcolsev}}
        1 & -30.06 & 86.09  & 123.97   & -106.86 & 48.58  & \textcolor{red}{nan}     & -21.32 \\
        2 & 149.94 & -86.09 & -56.03   & -106.86 & 48.58  & \textcolor{red}{nan}     & -21.32 \\
        3 & -90    & 0      & 115.96   & -106.86 & 131.42 & 150.52  & -21.32 \\
        4 & 90     & 0      & -64.04   & -106.86 & 131.42 & 150.52  & -21.32
       \end{array}$ 
    \\ 
    \noalign{\hrule width 8.5cm}
    HeLiu &
        $\begin{array}{p{\wcolzer}p{\wcolone}p{\wcoltwo}p{\wcolthr}p{\wcolfou}p{\wcolfiv}p{\wcolsix}p{\wcolsev}}
        1           & 90     & 0      & -64.04   & -106.86 & 131.42 & 150.52  & -21.32 \\
        2           & 90     & 0      & -64.04 & -106.86 & 131.42 & 150.52  & -21.32 \\
        3           & -30.06 & 86.09  & 123.97   & \textcolor{red}{nan}     & \textcolor{red}{nan}    & \textcolor{red}{nan}     & \textcolor{red}{nan} \\
        4           & 149.94 & -86.09 & -56.03   & \textcolor{red}{nan}     & \textcolor{red}{nan}    & \textcolor{red}{nan}     & \textcolor{red}{nan}
        \end{array}$   
    \\ 
    \noalign{\hrule width 8.5cm}
    IKFast &
        $\begin{array}{p{\wcolzer}p{\wcolone}p{\wcoltwo}p{\wcolthr}p{\wcolfou}p{\wcolfiv}p{\wcolsix}p{\wcolsev}}
        1           & -30.06 & 86.09  & 123.97   & -106.86 & 48.58  & \textcolor{red}{-137.3} & -21.32 \\
        2           & 149.94 & -86.09 & -56.03   & -106.86 & 48.58  & \textcolor{red}{-137.3} & -21.32
        \end{array}$ 
    \\
    \noalign{\hrule width 8.5cm}
    IKGeo  &
        $\begin{array}{p{\wcolzer}p{\wcolone}p{\wcoltwo}p{\wcolthr}p{\wcolfou}p{\wcolfiv}p{\wcolsix}p{\wcolsev}}
        1           & 149.93 & -86.09 & -56.03 & -106.9 & 48.58 & \textcolor{red}{-137.3} & -21.32 \\
        2           & -30.06 & 86.09 & 123.97 & -106.9 & 48.58  & \textcolor{red}{-137.3} & -21.32 \\
        3           & \textcolor{orange}{-27.9} & \textcolor{orange}{-0.02} & \textcolor{orange}{61.7559} & \textcolor{orange}{-106.9} & \textcolor{orange}{131.4} & \textcolor{orange}{150.5} & \textcolor{orange}{-21.32} \\
        4           & \textcolor{orange}{152.1} & \textcolor{orange}{0.02} & \textcolor{orange}{-118.2} & \textcolor{orange}{-106.9} & \textcolor{orange}{131.42} & \textcolor{orange}{150.5} & \textcolor{orange}{-21.32} \\
        5           & \textcolor{orange}{-174.8} & \textcolor{orange}{-48.83} & \textcolor{orange}{65.06} & \textcolor{orange}{53.35} & \textcolor{orange}{-90} & \textcolor{orange}{-173.4} & \textcolor{orange}{-21.32} \\
        6           & \textcolor{orange}{5.17} & \textcolor{orange}{48.83} & \textcolor{orange}{-114.9} & \textcolor{orange}{53.35} & \textcolor{orange}{-90} & \textcolor{orange}{-173.4} & \textcolor{orange}{-21.32}
        \end{array}$ 
    \\
    \noalign{\hrule width 8.5cm height 1.5pt}
    \multicolumn{2}{c}{\rule{0pt}{8pt} \textbf{Case 2 (algorithmic singularity, solvers with q6 as free variable)}} \\
    \noalign{\hrule width 8.5cm}
    {} & 
        $\begin{array}{p{\wcolzer}p{\wcolone}p{\wcoltwo}p{\wcolthr}p{\wcolfou}p{\wcolfiv}p{\wcolsix}p{\wcolsev}}
        sol & \ensuremath{q_1} & \ensuremath{q_2}  & \ensuremath{q_3} & \ensuremath{q_4} & \ensuremath{q_5} & \ensuremath{q_6} & \ensuremath{q_7} 
        \end{array}  $
    \\
    \noalign{\hrule width 8.5cm}
    GeoFIK (ours) & 
        $\begin{array}{p{\wcolzer}p{\wcolone}p{\wcoltwo}p{\wcolthr}p{\wcolfou}p{\wcolfiv}p{\wcolsix}p{\wcolsev}}
        1           & -18.99 & 78.13  & 28.75    & -39.33  & 0      & 204.90  & 0 \\
        2           & 161.01 & -78.13 & -151.25  & -39.33  & 0      & 204.90  & 0 \\
        3           & 2.37   & \textcolor{red}{nan}    & -148.49  & -39.33  & \textcolor{red}{nan}    & 169.24  & 0 \\
        4           & \textcolor{red}{nan}    & \textcolor{red}{nan}    & 31.51    & -39.33  & \textcolor{red}{nan}    & 169.24  & 0 \\
        5           & -12.43 & 90.22  & 28.08    & -14.19  & 0      & 193.44  & 0 \\
        6           & \textcolor{red}{nan}    & -90.22 & -151.92  & -14.19  & 0      & 193.44  & 0 \\
        7           & -4.97  & \textcolor{red}{nan}    & -151     & -14.19  & \textcolor{red}{nan}    & 180.65  & 0 \\
        8           & \textcolor{red}{nan}    & \textcolor{red}{nan}    & 29       & -14.19  & \textcolor{red}{nan}    & 180.65  & 0
        \end{array}$
    \\
    \noalign{\hrule width 8.5cm}
    IKFast & 
        no solution
    \\
    \noalign{\hrule width 8.5cm}
    IKGeo & 
        no solution
    \\
    \noalign{\hrule width 8.5cm height 1.5pt}
\end{tabular}
\end{scriptsize}
\end{table}

\section{Comparison with Baseline Solvers}\label{sec:comparison}

This section presents an analytical comparison against the state-of-the-art solvers from He and Liu \cite{he_liu} (referred to here as HeLiu), IKFast \cite{ikfast}, and the 6-DOF solver IKGeo \cite{ikgeo}. HeLiu is only compared fixing $q_7$ since it is the only solver they provide. For IKGeo, we coded the necessary functions to solve the IK of the Franka using the general 6-DOF solver. The different solvers provided by IKGeo were tested and it was decided to use {\small \texttt{IK\_spherical\_2\_intersect}} when locking $q_7$, and {\small \texttt{IK\_spherical}} for $q_6$ and $q_4$.


\subsection{Number of solutions attempted}

As discussed in Sections \ref{sec:sol_q7}-\ref{sec:sol_q4}, each case of GeoFIK solver attempts 8 solutions which constitute all geometrically valid solutions. 
In contrast, HeLiu attempts only 4 solutions, explicitly omitting a secondary elbow-down (solution 2 in Eq. (\ref{eq:alpha_2})), as acknowledged in \cite{he_liu}. 
Due to its general algebraic approach, IKFast attempts the 16 solutions that the system of equations can have. However, since only 8 solutions are geometrically possible, the other 8 must be complex or repeated. Among those, IKFast will return elbow-down solutions with link interference which have to be discarded by subsequent joint-limits checks.
The number of solutions attempted by IKGeo depends on each subproblem it solves. 



\definecolor{colorcell}{rgb}{0.851,0.890,0.820}

\begingroup

\begin{table*}[t]
\centering
\begin{scriptsize}

\caption{IK Solver Quantitative Comparison in terms of computation time, number of solutions, and errors.}\vspace{-5pt}
\label{IKComparison}
\begingroup
\setlength{\aboverulesep}{1pt}
\setlength{\belowrulesep}{1pt}
\begin{tabular}{cc @{\hspace{5pt}}c @{\hspace{6pt}}c @{\hspace{6pt}}c @{\hspace{6pt}}c @{\hspace{6pt}}c @{\hspace{6pt}}c @{\hspace{6pt}}c @{\hspace{6pt}}c @{\hspace{6pt}}c @{\hspace{3pt}}c}
    \toprule
    \multirow{2}{*}{\textcolor{black}{Min}/\textcolor{black}{Max}} &\multicolumn{4}{c}{Locking q7}         & \multicolumn{3}{c}{Locking q6} & \multicolumn{3}{c}{Locking q4} 
 & Locking S. Angle \\ \cmidrule(lr){2-5} \cmidrule(lr){6-8} \cmidrule(lr){9-11} \cmidrule(lr){12-12} 
	\textcolor{blue}{Mean} $\pm$  \textcolor{violet}{Std} & \textbf{GeoFIK}    & HeLiu & IKFast & IKGeo & \textbf{GeoFIK}    & IKFast  & IKGeo  & \textbf{GeoFIK}   & IKFast & IKGeo & \textbf{GeoFIK}    \\ 

\midrule

\multirow{2}{*}{}    

Joint angles
&  \cellcolor{colorcell} 1.7/4.8 
&   0.3/16.6 
&  3.2/14.6
&  4.5/25.0 
&  \cellcolor{colorcell} 2.5/6.3
&   33.7/129.3
&   7.7/30.1
&   \cellcolor{colorcell} 5.1/7.4
&    16.6/71.7
&    5.5/23.4
&    41.7/1429.2    \\

{computation time ($\mu\text{s}$)}
& \textcolor{blue}{2.1}$\pm$\textcolor{violet}{0.6} 
&  \cellcolor{colorcell} \textcolor{blue}{1.0}$\pm$\textcolor{violet}{0.6} 
&\textcolor{blue}{4.8}$\pm$\textcolor{violet}{1.7}
&\textcolor{blue}{6.0}$\pm$\textcolor{violet}{2.3} 
&  \cellcolor{colorcell} \textcolor{blue}{3.1}$\pm$\textcolor{violet}{0.9}  
&\textcolor{blue}{39.1}$\pm$\textcolor{violet}{7.8}
&\textcolor{blue}{8.3}$\pm$\textcolor{violet}{1.3}
&  \cellcolor{colorcell}\textcolor{blue}{5.4}$\pm$\textcolor{violet}{0.2}  
&\textcolor{blue}{18.7}$\pm$\textcolor{violet}{5.6} 
&\textcolor{blue}{6.1}$\pm$\textcolor{violet}{2.2} 
&\textcolor{blue}{198}$\pm$\textcolor{violet}{123} \\

\midrule

Jacobian only

&    \cellcolor{colorcell} 0.2/1.9
&   $-$
&   $-$
&   $-$
&   \cellcolor{colorcell}0.4/2.3
&   $-$
&   $-$
&   \cellcolor{colorcell} 0.9/7.7
&   $-$
&   $-$
&   30.8/1405.5
\\

computation time ($\mu\text{s}$)
&  \cellcolor{colorcell} \textcolor{blue}{0.3}$\pm$\textcolor{violet}{0.07} 
& {$-$}
& {$-$}
& {$-$}
&  \cellcolor{colorcell} \textcolor{blue}{0.8}$\pm$\textcolor{violet}{0.3}  
& \textcolor{black}{$-$} 
& \textcolor{black}{$-$} 
&  \cellcolor{colorcell} \textcolor{blue}{1.4}$\pm$\textcolor{violet}{0.5}
& \textcolor{black}{$-$}
& \textcolor{black}{$-$}
&\textcolor{blue}{192.1}$\pm$\textcolor{violet}{125}
\\
\midrule

Jacobian + joint angles   

&   \cellcolor{colorcell}  1.8/11.3
&   6/20
&  8.6/41
&  11.5/55.4 
&    \cellcolor{colorcell} 2.6/11.1
&   39.5/132.6
&   18/73.5
&    \cellcolor{colorcell} 5.2/21
&   17/65.5
&   10.5/31.6
&   41.6/1322.8
\\

computation time ($\mu\text{s}$)

&   \cellcolor{colorcell} \textcolor{blue}{2.7}$\pm$\textcolor{violet}{1.5} 
&\textcolor{blue}{7}$\pm$\textcolor{violet}{1.2} 
&\textcolor{blue}{12.8}$\pm$\textcolor{violet}{5} 
&\textcolor{blue}{14.4}$\pm$\textcolor{violet}{4.6} 
&   \cellcolor{colorcell} \textcolor{blue}{3.5}$\pm$\textcolor{violet}{1.3}  
&\textcolor{blue}{46.7}$\pm$\textcolor{violet}{7.8}  
&\textcolor{blue}{19.7}$\pm$\textcolor{violet}{3}  
&   \cellcolor{colorcell} \textcolor{blue}{5.6}$\pm$\textcolor{violet}{0.6}
&\textcolor{blue}{19}$\pm$\textcolor{violet}{4.6}
&\textcolor{blue}{11.4}$\pm$\textcolor{violet}{2.7}
&\textcolor{blue}{204.2}$\pm$\textcolor{violet}{135.3}
\\
\midrule

%
Num. valid sols.       
&   \cellcolor{colorcell} 1/8
&  0/4
&    \cellcolor{colorcell} 1/8
&    \cellcolor{colorcell} 1/8
&    \cellcolor{colorcell} 1/8
&    \cellcolor{colorcell} 1/8
&   0/8
&    \cellcolor{colorcell} 1/8
&    \cellcolor{colorcell} 1/8
&   0/4
&   0/8
\\

per call
&    \cellcolor{colorcell} \textcolor{blue}{3.2}$\pm$\textcolor{violet}{1.5}   
&\textcolor{blue}{2.6}$\pm$\textcolor{violet}{1.1}   
&   \cellcolor{colorcell}\textcolor{blue}{3.2}$\pm$\textcolor{violet}{1.5}
&   \cellcolor{colorcell}\textcolor{blue}{3.2}$\pm$\textcolor{violet}{1.5}
&   \cellcolor{colorcell} \textcolor{blue}{3.6}$\pm$\textcolor{violet}{1.5}  
&   \cellcolor{colorcell} \textcolor{blue}{3.6}$\pm$\textcolor{violet}{1.5}  
&\textcolor{blue}{2.9}$\pm$\textcolor{violet}{1.6}  
&   \cellcolor{colorcell}\textcolor{blue}{4.9}$\pm$\textcolor{violet}{2.0}   
&   \cellcolor{colorcell}\textcolor{blue}{4.9}$\pm$\textcolor{violet}{2.0} 
&\textcolor{blue}{2.6}$\pm$\textcolor{violet}{1.5} 
&\textcolor{blue}{2.0}$\pm$\textcolor{violet}{1.0}    \\

\midrule

Total num. of valid sols. 
&    \cellcolor{colorcell} 6352 
&   5272
&   6346
&   6350
&    \cellcolor{colorcell}7119
&    \cellcolor{colorcell}  7119
&   5856
&    \cellcolor{colorcell} 9742
&    \cellcolor{colorcell} 9742
&   5137
&   4042
\\

\midrule

Total num. of wrong sols. 
&    \cellcolor{colorcell} 0
&   72
&    \cellcolor{colorcell} 0
&   328
&   \cellcolor{colorcell} 0
&    \cellcolor{colorcell} 0
&   \cellcolor{colorcell}  0
&    \cellcolor{colorcell} 0
&   \cellcolor{colorcell}  0
&   \cellcolor{colorcell}  0
&   0
\\

\midrule




EE pos. error mean (m)
& \textcolor{blue}{$2 \mathrm{x}10^{-6}$} 
& \textcolor{blue}{$3 \mathrm{x}10^{-4}$}  
&   \cellcolor{colorcell} \textcolor{blue}{$7\mathrm{x}10^{-10}$}   
& \textcolor{blue}{$5 \mathrm{x}10^{-3}$}   
&   \cellcolor{colorcell}\textcolor{blue}{$9 \mathrm{x}10^{-14}$}   
& \textcolor{blue}{$7 \mathrm{x}10^{-10}$}    
& \textcolor{blue}{$9 \mathrm{x}10^{-8}$}  
&   \cellcolor{colorcell}\textcolor{blue}{$6 \mathrm{x}10^{-14}$}  
& \textcolor{blue}{$6 \mathrm{x}10^{-10}$}   
& \textcolor{blue}{$5 \mathrm{x}10^{-10}$}   
& \textcolor{blue}{$1 \mathrm{x}10^{-5}$}  
\\


\midrule
      


EE rot. error mean (rads)
& \textcolor{blue}{$1 \mathrm{x}10^{-5}$}  
& \textcolor{blue}{$4 \mathrm{x}10^{-4}$}  
& \textcolor{blue}{$1 \mathrm{x}10^{-9}$}
&  \cellcolor{colorcell} \textcolor{blue}{$3 \mathrm{x}10^{-13}$}
&  \cellcolor{colorcell} \textcolor{blue}{$3 \mathrm{x}10^{-13}$}   
& \textcolor{blue}{$1 \mathrm{x}10^{-9}$}
&   \cellcolor{colorcell} \textcolor{blue}{$3 \mathrm{x}10^{-13}$}
&  \cellcolor{colorcell} \textcolor{blue}{$3 \mathrm{x}10^{-13}$}  
& \textcolor{blue}{$1 \mathrm{x}10^{-9}$} 
&   \cellcolor{colorcell} \textcolor{blue}{$3 \mathrm{x}10^{-13}$} 
& \textcolor{blue}{$2 \mathrm{x}10^{-13}$}  
\\


  \bottomrule
	\end{tabular}
        \endgroup
 \end{scriptsize}
\end{table*}

\endgroup


\subsection{Handling of singularities}\label{sec:sing_comparison}

We examine the behavior of the solvers in the flat shoulder and wrist-on-$\mathbf{S}_7$ configurations described in Sec. \ref{sec:sing}.

\noindent{\bf flat shoulder singularity:} Consider the following IK problem that leads to a flat shoulder:
{\small%
\begin{align} 
\rotmat{O}{E} = & \begin{pmatrix}
0.6688331 & 0.31705344 & 0.672413 \\
-0.6398146 & -0.21507724 & 0.7378205 \\
 0.3785493 & -0.92369843 & 0.0590046
\end{pmatrix},
\nonumber\\
\pv{O}{r}{E/O} = &(0.61674948,0.32278029,0.56790512)^{\top},
\nonumber\\
q_7 = &  -21.32455095^{\circ}.\nonumber
\end{align}}%
The problem is sent to the solvers with $q_7$ as a free parameter. 
The results are shown as Case 1 in Table \ref{tab:sing}.

In GeoFIK, the user can provide a value of $q_1$ to resolve the redundancy. The default value is $\pi/2$. Solutions for the two ways of assembling the spherical shoulder are still returned with $\pm q_1$. One such solution is shown in Fig. \ref{fig:sing1}. Other two solutions that are not singular are returned, however, in those $q_6$ violates joint limits. In HeLiu, the user provides  a vector of joint angles from which it takes $q_1$ and uses it to resolve the redundancy induced by the singularity. Then, it 
returns two repeated solutions with $q_1$, rather than two different ones with $\pm q_1$, which can lead to discontinuity in the real-world applications, and the other two regular solutions that violate joint limits. IKFast fails in this singularity and only returns the two regular configurations that violate joint limits. IKGeo returns these same two regular configurations but also returns 4 invalid solutions for which the end-effector position is more than 1cm away from that of the requested pose.



{\bf shoulder-on-$\mathbf{S}_7$:} Consider the following IK problem that leads to an algorithmic singularity:
{\small%
\begin{align}
\rotmat{O}{E}= &\begin{pmatrix}
0.0746454 & -0.1964604 & 0.9776662\\
0.281646 & -0.93633263 & -0.2096583\\
0.9566105 & 0.2910058 & -0.0145606
\end{pmatrix},
\nonumber\\
\pv{O}{r}{E/O}= &(
0.89948341,
-0.1928922,
0.31960372)^{\top},
\nonumber\\
q_6= & 193.48937052^{\circ}.\nonumber
\end{align}}%
The problem is sent to the solvers with $q_6$ as a free parameter. 
The results are shown as Case 2 in Table \ref{tab:sing}. In the algorithmic singularities, $q_6$ stops being a free variable as there are only two values of $q_6$ that satisfy the pose. One of them is the value used in this IK problem.

GeoFIK is able to automatically detect that $q_6$ can no longer be freely chosen and transfers the problem to the solver with $q_7$ as a free variable. 8 solutions are found, indicating that triangle $SO_4O_6$ is almost flat, allowing elbow-up and elbow-down solutions as shown in Fig. \ref{fig:sing2}. Among those, two solutions are within the locked value for $q_6$. 

Both IKFast and IKGeo fail to generate any solution. It is worth highlighting that this behaviour is inconsistent in IKFast as it is able to find solutions to some shoulder-on-$\mathbf{S}_7$ singularities (yet, it is highly sensitive to numerical precision to work). This is not an issue with GeoFIK, since once it detects the singularity, it no longer requires the value of $q_6$. Hence, it returns consistently valid solutions.

\begin{figure*}[t]
\centering 
{\small
\def\svgwidth{0.93\linewidth}
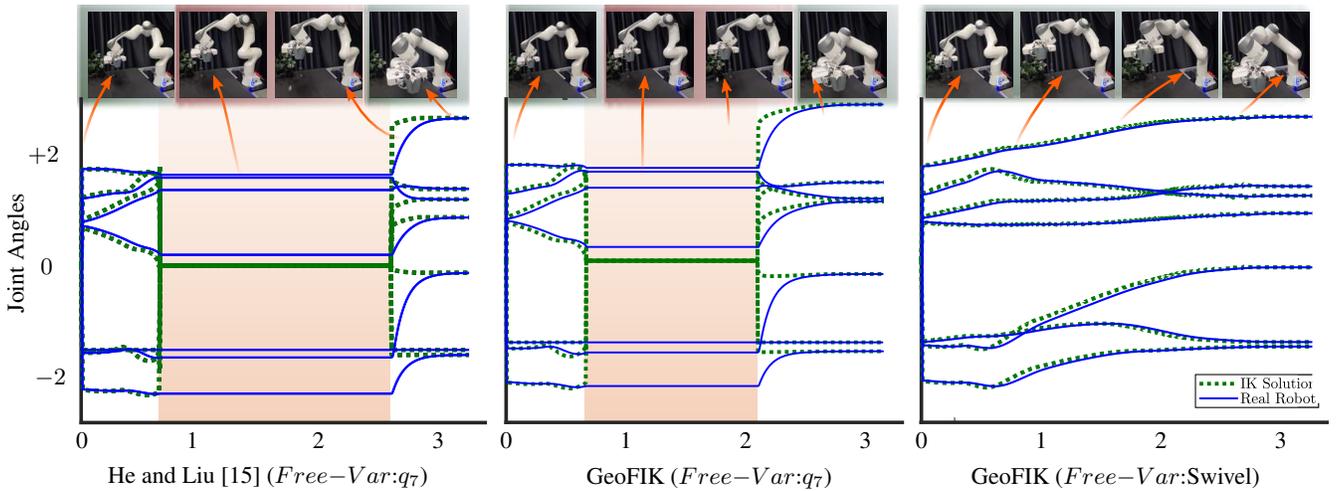
}
\caption{Trajectory tracking using IK solvers: The obtained joints (dashed-green lines) compared to the real-robot (blue lines). HeLiu and GeoFIK (with $q_7$ as a free-var) both fail to find a feasible solution around 10~s. (shaded area) and the robot stops (real-robot on the top strip) only to continue with an aggressive motion at the end of the trajectory. GeoFIK with the swivel-angle as free-var always finds a solution---as expected from Table~\ref{IKComparison}---enabling smooth joint-trajectory.}   
%
\label{fig::Lockingq7} 
\end{figure*}

\subsection{Quantitative Analysis}  

While Sec. \ref{sec:sing_comparison} provided a comparison in singular configurations, here we quantitatively evaluate the performance in regular and near-singular configurations.
We generated 2000 random valid joint configurations to produce EE poses, which in turn were sent as input to the solvers\footnote{Laptop used:  11th Gen Intel Core i7-11370H (3.30 GHz), 
32GB running Windows 11. Compiler: gcc 13.2.0 and optimization flag \texttt{-O3} for all solvers. For the case of locking $q_7$, clang was used instead for all solvers as it gave better results, except for IKFast which was compiled with gcc.}. 
%

%

We evaluated all the solvers against 4 parameters. The results are shown in Table~\ref{IKComparison}.%
 
\noindent\textbf{(i)} \textbf{Computational time} for a single call for only joint angles, only Jacobians, and both joint angles and Jacobians. For only joint angles, GeoFIK outperformed the other solvers when locking $q_6$ and $q_4$. HeLiu is the fastest in the case of $q_7$, this is due to the lack of calculation of elbow-down solutions.

When comparing the Jacobian computation time, GeoFIK outperformed all other solvers. This is mostly due to our screw-axis-based design that returns the Jacobian as a byproduct of the IK. In fact, GeoFIK outputs Jacobians even faster than the joint-configurations. 

\noindent\textbf{(ii)} \textbf{Number of distinct valid solutions (total and per call):} We consider a solution valid if it is within joint-limits and has an EE error smaller than 5mm in position and $3^{\circ}$ in rotation. Repeated solutions are not counted. It can be seen that GeoFIK outperforms all the solvers in total number of solutions with IKFast almost providing the same number. 

\noindent\textbf{(iii)} \textbf{Total number of incorrect solutions:} Incorrect solutions are defined here as being within joint-limits but having an EE error as described in (ii). HeLiu and IKGeo returned incorrect solutions in the $q_7$ case.

\noindent\textbf{(iv)} \textbf{EE error:} GeoFIK outperforms all the solvers in EE error with exception of the $q_7$ case.

\section{Real-World Experiment}\label{sec:experiments}

We evaluated our GeoFIK approach on a real Franka Robot to investigate (i) feasibility and joint-space tracking of a desired task-space trajectory (herein obtained through vector fields) within the robot’s central workspace, and (ii) whole-body tasks that require both the end-effector pose and the manipulability ellipsoid to match a human-demonstrated reference (i.e., sweeping and pushing). 

\textbf{Feasibility and Trajectory Tracking:} This first scenario tested the ability to compute real-time IK solutions and maintain a continuous joint-space trajectory over a basic trajectory in the central region of the workspace, a domain expected to be straightforward. Despite this assumption, the IK solver by HeLiu failed after approximately 10 s, providing no valid solutions and forcing the robot to halt. Our GeoFIK solver, when using $q_7$ as the free parameter, also reached a point where it could not find a solution. Both failures led to abrupt reconfiguration once a valid solution was found again (Fig.~\ref{fig::Lockingq7} top-images with a red border). 

\begin{figure}[t]
\centering
{\footnotesize   
\def\svgwidth{0.98\columnwidth}
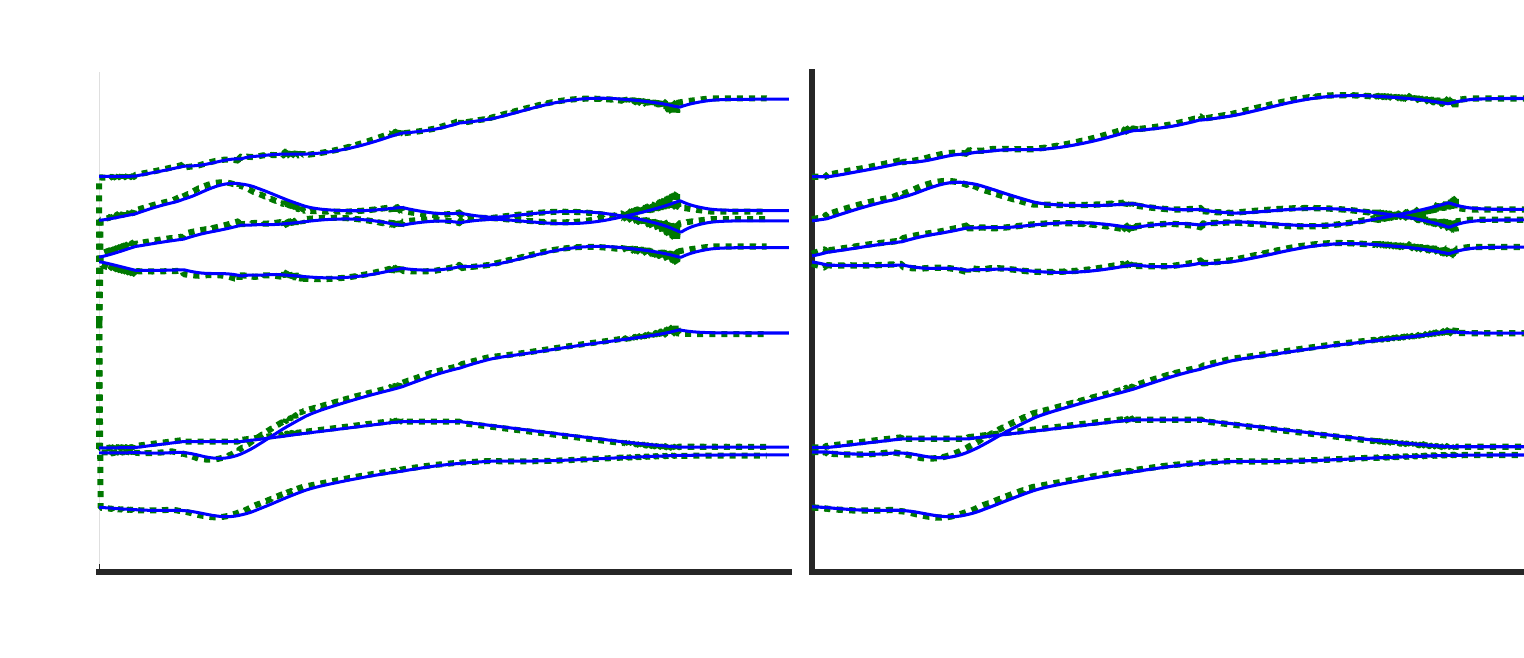
}
\caption{Trajectory tracking using IK solvers and sampling $q_7 \pm 10^{o}$. Sampling allows $q_7$-locking IKs to complete the task, but the resulting trajectory includes chattering due to discontinuities (the overlapped green areas).
}   
\label{sampling} 
\end{figure}

\begin{figure}[t]
\begin{subfigure}{0.3\linewidth}
		\centering
		\includegraphics[width=1\linewidth,height=0.9\linewidth]{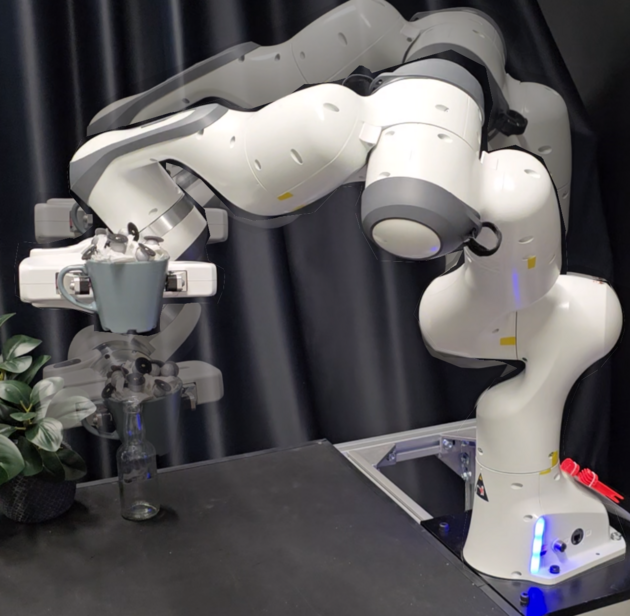}
		\caption{IKFast EE Track}
		\label{fig::IKFatsEETrack}
	\end{subfigure}
 \begin{subfigure}{0.3\linewidth}
		\centering
		\includegraphics[width=1\linewidth,height=0.9\linewidth]{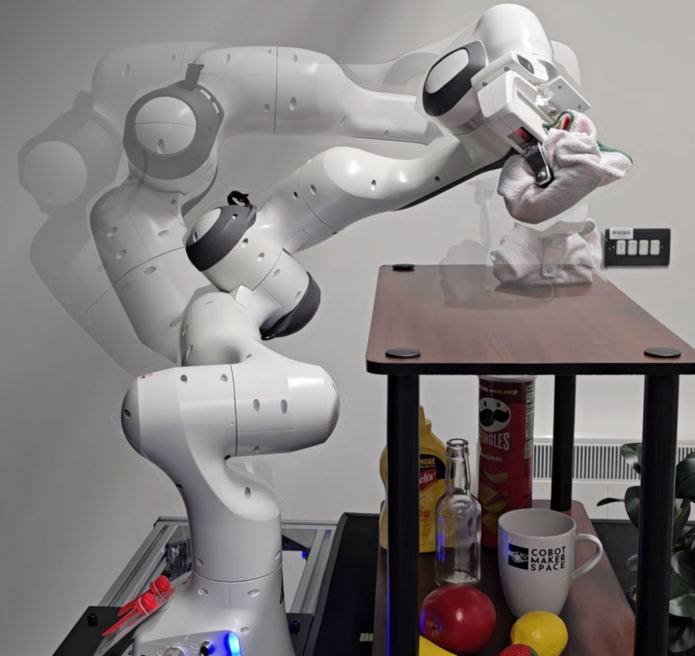}
		\caption{IKFast Sweep}
		\label{fig::IKFatstSweep}
	\end{subfigure}
     \begin{subfigure}{0.3\linewidth}
		\centering
		\includegraphics[width=1\linewidth,height=0.9\linewidth]{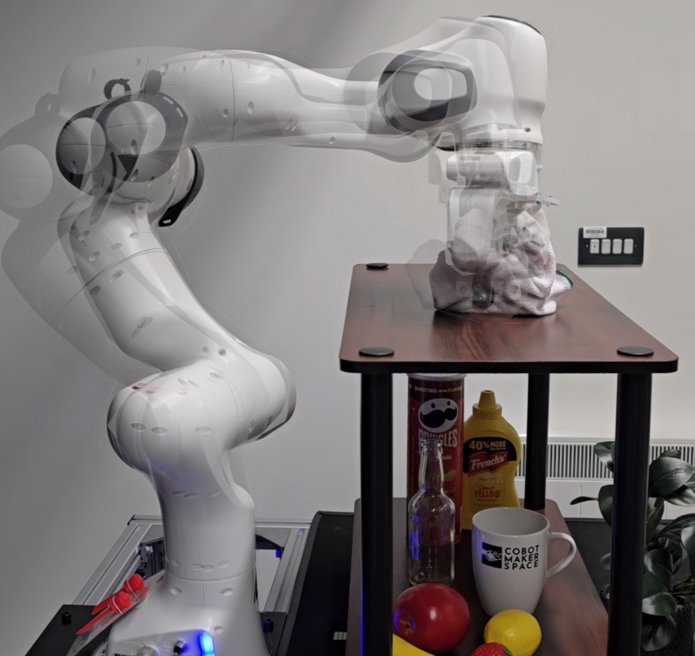}
		\caption{GeoFIK Sweep}
		\label{fig::GeoFIKSweep}
	\end{subfigure}
\caption{(a) IKFast traj. tracking, and (b)c IKFast and (c) GeoFIK sweeping tasks. IKFast often provides invalid solutions. In (a), the lack of solutions near the $\textbf{q} (0)$ results in abrupt arm rotations exceeding torque reflexes. Similarly, in sweeping tasks, IKFast causes the arm to flip over lacking  proper solutions. GeoFIK success in all tasks.  
}
\label{fig:IKFats}
\end{figure}

Allowing the standard $q_7$-locking solvers to sample a small range $\pm 10 ^{o}$ around the current  $q_7$ mitigated total failure, but induced frequent chattering in joint angles (Fig.~\ref{sampling}). Meanwhile, IKFast remained unable to find even a valid initial solution, all valid IKs demand an instantaneous flip of the base joints, generating torque reflexes and halting progress altogether, as shown in Fig.~\ref{fig:IKFats}(a). In contrast, the swivel-angle-based GeoFIK solution never lost feasibility and preserved a smooth trajectory as shown in Fig~\ref{fig::Lockingq7}.

\noindent \textbf{Whole-Body (Manipulability) Tracking:}  
The second set of experiments required each solver to produce both joint angles and Jacobian, which was used to compute the manipulability ellipsoid. The solver then aimed to match a reference ellipsoid provided by a user demonstration (e.g. sweeping or pushing motions). We sampled 250 configurations every 30ms, akin to a lightweight model-predictive control loop.  

%

\begin{figure}[h]
\centering{\small%
        \def\svgwidth{0.44\textwidth}%
        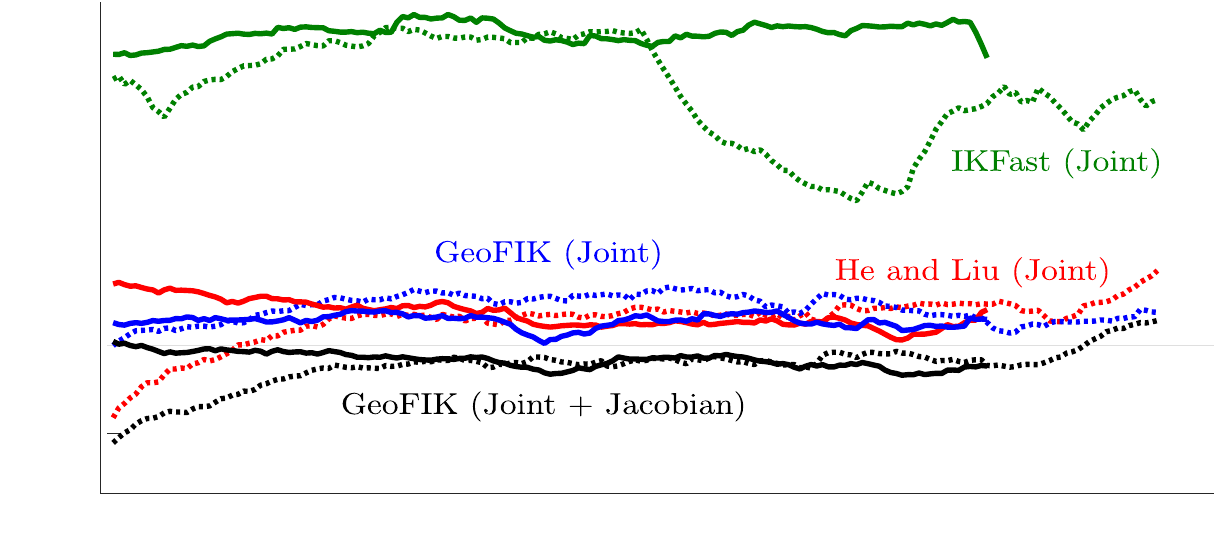%
        } 
        \caption{Computation time in tracking manipulability for sweeping (dashed) and pushing (solid) tasks. 
        }
\label{fig:computationTimeSweepingPushing}
\end{figure}

As shown in Fig.~\ref{fig:computationTimeSweepingPushing}, GeoFIK achieves lower computational times thanks to its screw-transformation design, where the Jacobian emerges naturally from the IK solution. Although HeLiu managed to complete the sweeping task, it proved slower in delivering joint–Jacobian pairs. IKFast again frequently produced no valid solutions (or an out-of-reach configuration), leading to task failure in the sweeping experiments, Figs.~\ref{fig:IKFats}(b-c). GeoFIK, in contrast, consistently offered feasible joint trajectories while preserving the desired manipulability profile, underscoring the benefits of its geometry-aware method and multi-parameter redundancy resolution.

\section{Conclusions}\label{sec:conclusions}

We introduced a fully geometric, screw-theory-based IK solver for the Franka Robot, accommodating multiple free parameters and automatically providing the Jacobian even faster than the joint configuration. By systematically capturing all geometrically valid solutions, our method, GeoFIK, overcomes the limitations of existing solvers that either fix only $q_7$ or neglect key geometric insights. Extensive quantitative benchmarks showed that our approach consistently outperforms baseline solvers in terms of reliability, completeness, and efficiency. Real-world experiments confirmed its resilience, enabling the robot to smoothly execute tasks and even when controlling less conventional variables, such as the swivel angle. We believe that the flexibility and robustness of GeoFIK can facilitate more advanced planning,  control, and learning algorithms, thereby fulfilling the growing demand for speed and reliability in modern robotics.

\normalsize
\bibliography{biblio}


\end{document}